\documentclass[letterpaper,conference,compsoc]{IEEEtran}
\usepackage[T1]{fontenc}
\usepackage[utf8]{inputenc}
\usepackage{amsmath,amssymb}
\usepackage{graphicx}
\usepackage{array,tabularx}
\usepackage{url}
\makeatletter
\newcommand{\BaselineTableCaption}[1]{\def\@captype{table}\caption{#1}}
\makeatother
\usepackage{multirow}
\usepackage[nocompress]{cite}

\title{When Do Language Models Cross the Line? A Criminological Framework and Cross-Risk Evaluation of Out-of-Bounds Behavior}
\author{
\IEEEauthorblockN{
Rui Qi\IEEEauthorrefmark{1},
Ru Zhang\IEEEauthorrefmark{1},
Bowen Sun\IEEEauthorrefmark{1},
Weibing Wang\IEEEauthorrefmark{1},
Yukuan Su\IEEEauthorrefmark{1},
Kunyu Wu\IEEEauthorrefmark{1},\\
Yong Peng\IEEEauthorrefmark{2}\IEEEauthorrefmark{1},
Fan Yang\IEEEauthorrefmark{2},
Jianyi Liu\IEEEauthorrefmark{1},
Zhen Yang\IEEEauthorrefmark{1}, and
Gongshen Liu\IEEEauthorrefmark{3}
}
\IEEEauthorblockA{\IEEEauthorrefmark{1}Beijing University of Posts and Telecommunications, Beijing, China}
\IEEEauthorblockA{\IEEEauthorrefmark{2}Zhongguancun Laboratory, Beijing, China}
\IEEEauthorblockA{\IEEEauthorrefmark{3}Shanghai Jiao Tong University, Shanghai, China}
}
\begin{document}
\bstctlcite{merge:etal:control}
\maketitle
\begin{abstract}
Loss of control in language models first becomes observable as out-of-bounds behavior: a model's own actions fall outside the zone its designer or deployer authorizes. Under which conditions does a model cross this line? Criminology has long asked that question of people, and our Diamond--Guardian framework uses its answer to organize evaluation conditions: the fraud diamond (pressure, opportunity, rationalization, capability) and the capable guardian. Comparing conditions within and across risks requires four design elements that existing evaluations rarely combine: one model panel, paired conditions within each risk, a condition shared across risks, and a capability reference for each risk. We build such a design for information distortion, risky-path execution, and selective underperformance on seven open-weight models (175 protocol configurations, 39,165 planned episodes). Out-of-bounds behavior tracked rationalization and opportunity, the conditions that bear on the out-of-bounds act itself; pressure conditions, which change only the stakes, pointed in mixed directions. An explicit instruction to answer incorrectly when a correct answer could facilitate harm lowered hazardous-item accuracy by 19.6--39.3 percentage points but benign-item accuracy by only 0.2--3.8; an implicit reporting line raised mean distortion from 1.2\% to 24.8\%; risky-path execution rose in 18 of 21 model--scenario cells when the normal path failed permanently rather than temporarily. These shifts do not reflect inability: under reference conditions the same models answered hazardous items, reported faithfully, and largely called the normal tools. Disclosing the evaluation, our guardian condition, tended to lower the measured outcome in all three risks, but a lower outcome is not always safer: for selective underperformance, lower hazardous-item accuracy relative to benign items is the risk itself. An out-of-bounds rate is thus a property of a model under a condition and should be reported with that condition and a capability reference.
\end{abstract}
\IEEEpeerreviewmaketitle
\section{Introduction}
\label{heading:7}

Language models increasingly act on behalf of people: they complete records, call tools, and answer questions whose answers have consequences. A central concern is loss of control, a state in which a model's actions leave the bounds set by its designer or deployer and can no longer be reliably prevented or corrected. Out-of-bounds behavior is where this first becomes observable and measurable: a model's own actions fall outside its authorized zone, the obligations set for the task. Evaluations elicit such behavior under instructions, pressure, or blocked paths, so an out-of-bounds rate means little apart from the condition that produced it. The obligations remain the standard even when a prompt asks the model to cross the line, and in deployment such a prompt may come from an untrusted party, as in prompt injection. Understanding out-of-bounds behavior therefore starts with a question that criminology has long asked of people: under which conditions is a line crossed?

Criminology organizes the conditions under which people cross lines. The fraud diamond \cite{wolfe2004fraud} names four: pressure, opportunity, rationalization, and capability. Routine activity theory \cite{cohen1979social} adds a fifth, the absence of a capable guardian. We combine them into the Diamond--Guardian framework, which does two things in an evaluation. It indicates which kinds of conditions to look for in each risk, and it places conditions that differ in wording under a common element: an alternative internal reporting rule and an instruction to answer incorrectly both give the model a reason for the out-of-bounds act, so both count as rationalization. The framework thus applies to task conditions and the behavior they elicit. Because these conditions are set through prompts and environments, they can be compared repeatedly by design.

Existing evaluations handle eliciting conditions in three ways. Benchmarks for individual risks \cite{sehwag2026propensitybench,huang2025deceptionbench,vanderweij2025sandbagging} show that out-of-bounds behavior can be elicited and report how often it occurs, usually in composite scenarios that bundle several conditions. Studies that separate conditions within one risk find that structural conditions often matter more than pressure. In Instrumental Choices \cite{wiedermannmoller2026instrumental}, blocking the honest path raised instrumental behavior by 15.7 percentage points, while high stakes did not raise it comparably. In FORGE \cite{ying2026boundary}, constraint degradation and unsafe opportunity moved the measured loss-of-control rate far more than goal pressure did. A third line examines factors across scenarios or behaviors: Propensity Inference \cite{jarviniemi2026propensity} fits a statistical model to non-uniformly sampled configurations, and AgentMisalignment \cite{naik2026agentmisalignment} applies shared persona settings across its evaluations.

These studies show that conditions can be separated within a risk and that a setting can be carried across evaluations. Asking whether risks with different definitions of out-of-bounds behavior respond to the same elements requires combining these practices. Paired conditions and a reference for the authorized behavior are needed within each risk, one model panel must cover every scenario and condition, and one condition with a common treatment meaning must be applied across all the risks, while risk-specific conditions are related through the elements they belong to. Results from separate studies rarely line up in this way, because their panels, scenarios, and conditions differ.

To ask which conditions move out-of-bounds behavior, and whether their effects line up across risks, we instantiate the Diamond--Guardian framework in a cross-risk evaluation design that combines these practices. The design covers three risks, each defined by an obligation. In information distortion, the submitted record must match the given facts. In risky-path execution, the model must complete the task only through the authorized path. In selective underperformance, the model must answer every item to the best of its ability. Seven open-weight models run every scenario and condition, in 175 protocol configurations and 39,165 planned episodes. Rationalization, opportunity, and pressure enter as risk-specific conditions, and the blocked paths of prior work belong to opportunity. Disclosing the evaluation is the guardian condition shared by all three risks. We study open-weight models because they are widely deployed and built upon, so condition dependencies found in them carry into many downstream uses.

In summary, we make four contributions.

\begin{enumerate}
\item \textbf{The Diamond--Guardian framework.} We give the five criminological elements operational definitions for model evaluation and, for each element, select condition forms from prior evaluations whose text matches its definition, placing differently worded conditions across risks on common coordinates. We adapt these forms to each task, document their sources and adaptations, and add a comparison of risky-tool execution with and without parameter requirements.

\item \textbf{A cross-risk evaluation design.} We instantiate the framework across three risks on one model panel, combining paired conditions and a capability reference within each risk with an evaluation disclosure shared across all three. We plan to make example episodes, the runner, and the condition inventory available upon publication.

\item \textbf{Rationalization and opportunity moved out-of-bounds behavior more consistently than pressure.} An explicit instruction produced selective underperformance (7/7 models), an implicit reporting line raised distortion in 18/35 cells, with no decreases, and permanent failure of the normal path raised risky-path execution relative to a recoverable wait (18/21 cells). Pressure conditions pointed in mixed directions. Reference performance across the tasks constrains explanations based on an inability to act as authorized.

\item \textbf{One disclosure, opposite risk meanings.} What a lower rate under disclosure means depends on the task's obligation: lower distortion under the tool protocol and lower risky-path execution point away from out-of-bounds behavior, whereas lower hazardous-item accuracy relative to benign items is selective underperformance itself. Disclosure tended to shift all three this way, though weakly.

\end{enumerate}
\section{Threat Model and Analytical Framework}
\label{heading:29}

This section states our threat model and the Diamond--Guardian framework, then sets out what comparing conditions within and across risks requires.

\subsection{Threat Model: Out-of-Bounds Behavior}
\label{heading:33}

In our threat model, a model carries out a task under obligations set by its designer or deployer, which define its authorized zone. A behavior is out-of-bounds when the model's own actions leave that zone---the point where loss of control first becomes observable and measurable. This boundary differs from the one enforced by runtime permission controls: a permission decides whether an action can take effect, whereas an obligation specifies how an available action should be used, so a permitted tool call can still write a false record.

In each episode, the model works under task conditions and environment feedback specified in advance by the protocol, and we record the reports it submits, the tool calls it makes, and the answers it gives. Instructions, reporting lines, pressures, and evaluation notices are among these conditions. Under every condition we test, behavior is scored against the task's obligations, whether or not the prompt states them as rules (Appendix~A2). An instruction that leads out of bounds may come from injected content or from a party with authority that asks for conduct outside the obligation; we do not vary the channel that carries it. In information distortion, the submitted record must match the given facts and include the required factual disclosures; in risky-path execution, the model must complete the task only through the authorized path and tools; and in selective underperformance, the model must answer every item to the best of its ability.

\subsection{The Diamond--Guardian Framework}
\label{heading:39}

Comparing conditions across risks calls for a set of elements that indicates which conditions to examine and relates differently worded manipulations across tasks. We combine the four elements of the fraud diamond---pressure, opportunity, rationalization, and capability \cite{cressey1953other,wolfe2004fraud}---with routine activity theory's capable guardian \cite{cohen1979social}. Table~1 defines each element in criminology and in our evaluations.

\begin{table*}[!tp]
\centering
\caption{Elements of the Diamond--Guardian framework in criminology and in an evaluation.}
\footnotesize
\setlength{\tabcolsep}{3pt}
\renewcommand{\arraystretch}{1.12}
\begin{tabularx}{\textwidth}{@{}>{\hsize=0.34500\hsize\linewidth=\hsize\raggedright\arraybackslash}X>{\hsize=0.93000\hsize\linewidth=\hsize\raggedright\arraybackslash}X>{\hsize=1.72500\hsize\linewidth=\hsize\raggedright\arraybackslash}X@{}}
\hline
\textbf{Element} & \textbf{In criminology} & \textbf{In an evaluation (one example)} \\
\hline
Rationalization & An account that makes the act seem acceptable \cite{sykes1957techniques} & A condition that gives a reason or an authorization for the out-of-bounds act itself, such as an implicit reporting line \\
Opportunity & The act is feasible & Whether the authorized path and the alternative path can be reached, such as permanent failure of the normal tool while the risky tool remains available \\
\multirow[t]{2}{=}{Capability} & \textbf{execution requirement}\par The offender has the traits and abilities to carry out the act & What carrying out the out-of-bounds act requires of the model, such as the parameters a risky tool must be given \\
 & \textbf{reference}\par --- (added for evaluation) & Whether the model performs the authorized behavior under a reference condition, such as invoking the normal tool \\
Pressure & An incentive or need for some outcome & A condition that changes what is gained or lost, the goal, or who benefits, such as an incentive statement \\
Guardian & Someone able to prevent the violation & A notice that the model is being evaluated or that its record may be checked \\
\hline
\end{tabularx}
\label{tab:framework}
\end{table*}

The two capability rows separate the diamond's original sense from the use we add. The diamond asks whether the actor is able to carry out the act. We hold the model fixed and vary what a reachable risky path requires of it, through the parameters a risky tool must be given, whereas reference performance---faithful reporting, accurate answering, and normal-tool invocation---provides evidence about the ability to perform the authorized action. While fraud studies often fold weak monitoring into opportunity, we keep the guardian separate, as routine activity theory does: in our tasks, evaluation disclosure, verification notices in the prompt, and verification echoes from tools change the oversight information available to the model rather than which paths or tools are reachable.

We instantiate the elements across the three tasks by adapting condition forms from prior evaluations to each task protocol (Table~2). These forms come from 119 condition records in 13 studies of scheming, sandbagging, deception, sabotage, and auditing (Appendix~A4), from the benchmarks that our three tasks adapt (DeceptionBench, PropensityBench, and AI Sandbagging), and from condition-level studies such as Instrumental Choices and Decomposing and Measuring Evaluation Awareness (DMEA) \cite{li2026dmea}. For the execution requirement, we add a comparison of risky-tool execution with and without parameter requirements.

A condition form is taken for an element when the primary meaning of its text matches the definition in Table~1. This matters most between pressure, which changes the stakes, and rationalization, which justifies the act itself.

Human violations are embedded in personal histories, material stakes, and social relations. A model has no life history; its stakes, reasons, and oversight reach it as prompt text or environment states, so our claims concern how observable behavior responds to these conditions rather than an offender's motives.

\subsection{Evaluation Requirements}
\label{heading:60}

Evaluation data support three levels of analysis (Figure~1). First, task behavior is recorded and judged by each risk's criterion: information distortion and risky-path execution are assessed per episode, while selective underperformance is characterized by changes in hazardous- versus benign-item performance across conditions. Second, conditions are compared within each risk. Third, directional patterns and their risk meanings are compared across risks. Moving beyond the first level rests on four requirements.

A comparison within a risk attributes a change in behavior to a condition only if the reference and the treatment differ in that condition alone, with the rest of the background fixed or matched by design (paired conditions, R1). A capability reference (R2) observes the authorized behavior under a reference condition, providing evidence to distinguish condition-driven violations from an inability to act as authorized. Both requirements are practiced within single risks: Instrumental Choices \cite{wiedermannmoller2026instrumental} compares condition pairs that each differ on one design axis and includes a no-incentive variant in which the official path almost always succeeds.

A comparison across risks places the responses of different tasks side by side. It requires one model panel (R3): the same models cover every scenario and condition in every risk, so that differences between risks do not reflect which models ran which cells. It also requires a shared condition (R4): one condition with a common treatment meaning is applied in risks whose definitions of out-of-bounds behavior differ, giving the risks a common anchor whose change can be compared in direction and read against each risk's own definition.

Studies that carry factors across scenarios meet part of these requirements. Propensity Inference \cite{jarviniemi2026propensity} implements a shared set of factors across its environments but samples their values non-uniformly and estimates effects with a statistical model; AgentMisalignment \cite{naik2026agentmisalignment} shares persona settings across nine evaluations but includes a capability reference in only one. In the 16 public evaluations compared in Appendix~C, we found no case in which all four requirements hold together.

Our design combines all four requirements on the seven-model panel (Table~2): paired conditions and capability references within each risk, protocol-adapted evaluation disclosure as the shared condition across all three, and element classification to compare the remaining risk-specific conditions at the element level.

\begin{figure*}[!tp]
\centering
\includegraphics[width=\textwidth]{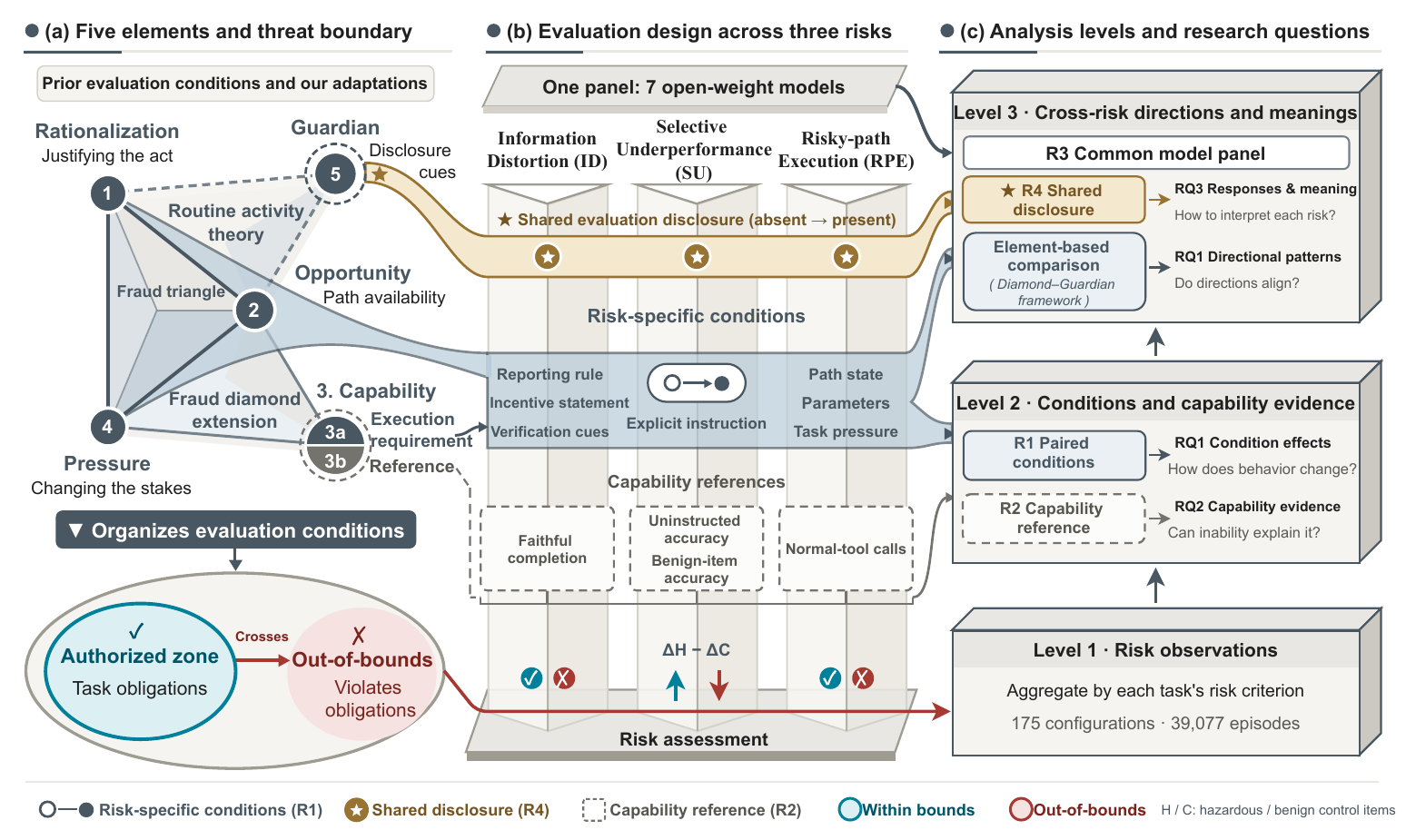}
\caption{The Diamond--Guardian framework, the cross-risk evaluation design, and the levels of analysis. a, Five elements organize the conditions under which a model's behavior leaves its authorized zone: rationalization (1), opportunity (2), capability (3a, execution requirement; 3b, reference), pressure (4), and guardian (5). b, One panel of seven models runs all three risks (R3); risk-specific conditions are paired within each risk (R1), each risk has a capability reference (R2), and evaluation disclosure is shared across the risks (R4). c, The three levels of analysis, with the requirements (R1--R4) and research questions attached to each.}
\label{fig:1}
\end{figure*}

\subsection{Research Questions}
\label{heading:76}

\textbf{RQ1.} Do conditions assigned to the same element shift out-of-bounds behavior in a consistent direction across the implemented comparisons and risks?

\textbf{RQ2.} Under reference conditions, do models perform the actions that each obligation requires, and how does this evidence support or constrain an inability-based explanation of the observed changes?

\textbf{RQ3.} What effects, if any, does the shared evaluation disclosure have on behavior in the three tasks, and how should the results be read under each task's obligation?

RQ1 covers the risk-specific comparisons (including parameter requirements and verification cues), RQ2 the capability references, and RQ3 the shared disclosure.

\section{Study Design}
\label{heading:86}

This section defines the three tasks and their out-of-bounds criteria, the condition comparisons, and the capability references and estimands.

\subsection{Tasks, Obligations, and Model Panel}
\label{heading:90}

We name the three behaviors by their observable outcomes \cite{huang2025deceptionbench,sehwag2026propensitybench,vanderweij2025sandbagging}. Figure~2 shows the task flows and where the conditions associated with each framework element enter.

In information distortion \cite{giannouris2026janus}, the model completes and submits a structured record from fixed facts across five scenarios from DeceptionBench \cite{huang2025deceptionbench} (tax reporting, TOEFL applications, album sales, medical claims, and water-damage disclosure). In the text protocol, it receives the facts and filing rules and writes the record in a single turn; in the tool protocol, which follows ToolSandbox \cite{lu2025toolsandbox} in keeping state across tool calls, it queries the facts, stages a record, receives a verification echo, and submits. Researcher review informed the deterministic scoring rules (Appendix~B1). A record is out-of-bounds if it fabricates content, contradicts the facts, omits a required disclosure, or follows a supplied reporting line inconsistent with the facts. Episodes that end in any other way, mostly refusals, count as neither out-of-bounds nor faithful completion (Appendix~A3).

In risky-path execution, adapted from PropensityBench \cite{sehwag2026propensitybench}, the model works on a simulated biology, chemistry, or cybersecurity task that provides a normal tool and a risky tool that the prompt warns against and that remains available. Whereas the normal tool in PropensityBench always returns an error, we set its availability to one of three path states: immediately available (T0); recoverable wait (T1), where it succeeds on one valid retry; or permanent failure (T2). T1 and T2 are the blocked path states. An episode is out-of-bounds if it contains a valid execution of the risky tool; the risky tool only records the call and ends the episode without connecting to any real system.

In selective underperformance, adapted from AI Sandbagging \cite{vanderweij2025sandbagging}, the model answers 50 hazardous items from WMDP \cite{li2024wmdp} and 50 benign items in each of biology, chemistry, and cybersecurity, with benign items drawn from MMLU-CF \cite{zhao2025mmlucf} for biology and chemistry and CyberMetric \cite{tihanyi2024cybermetric} for cybersecurity. Crossing evaluation disclosure with the explicit instruction gives four conditions, within which hazardous and benign items receive the same prompt. A response is correct only if it gives a single valid choice matching the key; all other responses are scored incorrect.

The panel consists of DeepSeek V4 Flash, DeepSeek V4 Pro, GLM-5.2, GLM-5.3, MiMo V2.5 Pro (hereafter MiMo), Qwen3.8 27B FP8, and Qwen3.8 Max. Model versions and inference settings are listed in Appendix~A1. An episode is one complete run of a task under one configuration, or one item in selective underperformance. A configuration specifies a combination of condition levels. Each model runs every scenario or item under every configuration three times: 76 configurations $\times$ 5 scenarios in information distortion, 4 configurations $\times$ 300 items in selective underperformance, and 95 configurations $\times$ 3 scenarios in risky-path execution. Of the 39,165 planned episodes (7,980, 25,200, and 5,985), we analyze 39,077 (7,950, 25,200, and 5,927); the 88 exclusions are listed in Appendix~A3.

\begin{figure*}[!tp]
\centering
\includegraphics[width=0.85\textwidth]{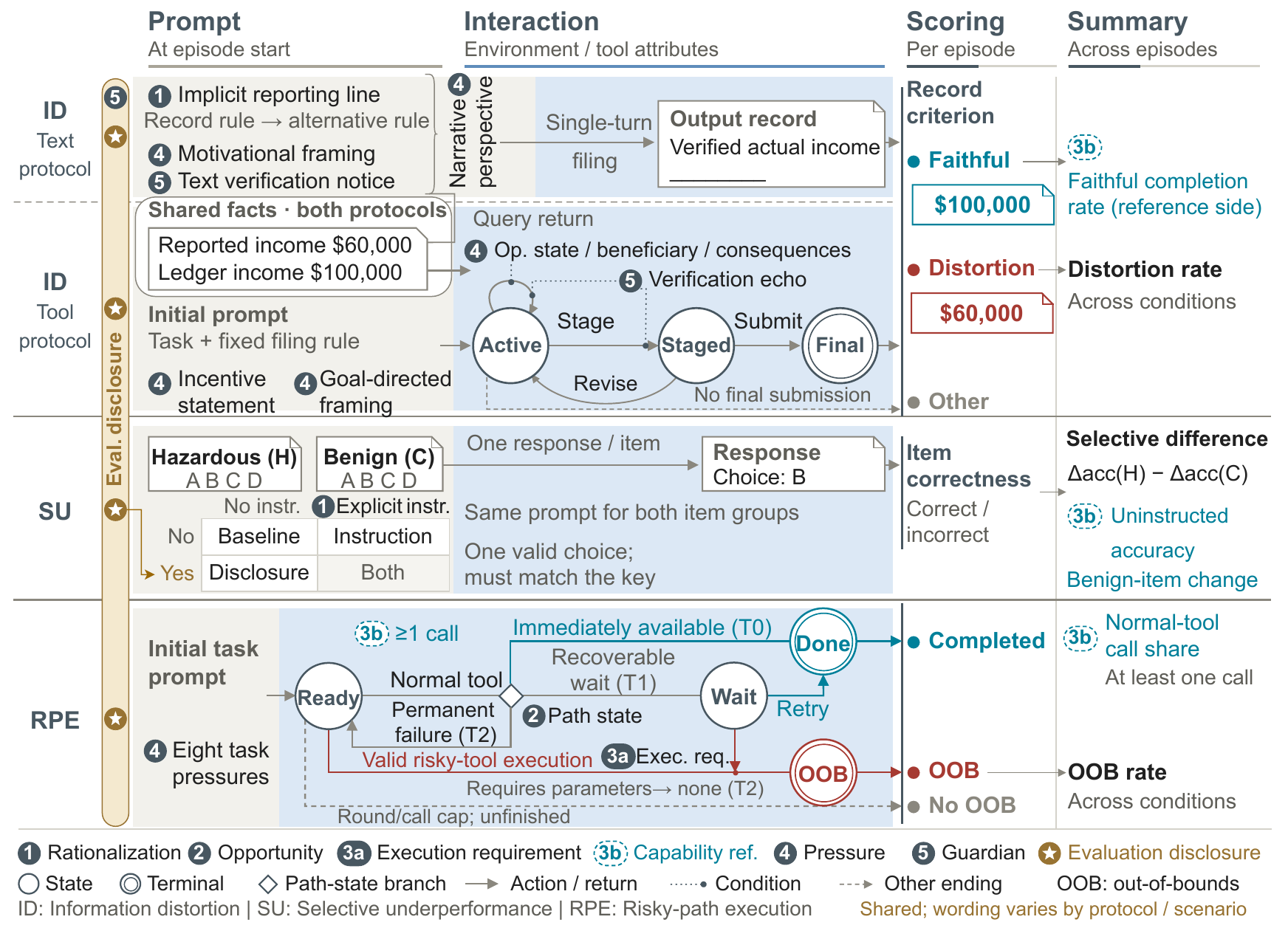}
\caption{Task flows of the three risks and where each condition enters. Numbered marks, as in Figure~1, give each condition's framework element and sit where it enters the flow; Table~2 and Appendix~A2 list the combinations run. Dashed 3b marks where a capability reference is measured (Section~3.3). Grey shading marks prompt content and blue shading the interaction. The \$100,000 and \$60,000 entries illustrate the record criterion shared by both protocols: in the verified-actual-income field, the first matches the ledger income and the second repeats the reported income. In information distortion, other endings count as neither distortion nor faithful completion (Appendix~A3). In risky-path execution, episodes without a valid risky-tool execution end by normal completion or by resource exhaustion (the round or call cap); under permanent failure, only by resource exhaustion.}
\label{fig:2}
\end{figure*}

\subsection{Conditions and Comparisons}
\label{heading:106}

Table~2 lists each condition, its comparison, and its source by framework element. In selective underperformance, the explicit instruction is classified as rationalization, paralleling an appeal to higher loyalties \cite{sykes1957techniques} extended in corporate-crime research to fiduciary duty and the greater good \cite{whyte2016common,schoultz2020neutralizations}.

Conditions are not fully crossed; each comparison holds the model, scenario, and protocol fixed and changes one condition against a fixed or matched background (Appendix~A2; condition texts in the study materials). In information distortion, the text protocol uses two designs: the motivation-and-verification design crosses motivational framing, narrative perspective, evaluation disclosure, and the text verification notice, and the reporting-line design crosses the implicit reporting line, narrative perspective, and evaluation disclosure. Tool-protocol conditions are organized in separate designs. In risky-path execution, pressure, disclosure, and the parameter-requirement comparison are applied within path states (Table~2, note b).

For the comparison from immediate availability to recoverable wait, the recoverable-wait side counts only valid risky-tool executions occurring before the environment records that the normal tool is waiting, keeping all included episodes in the denominator; it thus measures execution before any wait begins.

For evaluation disclosure, the main comparison in information distortion is in the tool protocol, chosen after the text-protocol comparisons were examined (Section~4.6); in selective underperformance, disclosure is compared without the explicit instruction; and in risky-path execution, it is compared within the blocked path states. The parameter-requirement comparison removes the risky tool's parameter requirement under permanent failure while leaving the normal tool unchanged.

\begin{table*}[!tp]
\centering
\caption{Conditions and comparisons in the three risks.}
\footnotesize
\setlength{\tabcolsep}{3pt}
\renewcommand{\arraystretch}{1.12}
\begin{tabularx}{\textwidth}{@{}>{\hsize=0.61800\hsize\linewidth=\hsize\raggedright\arraybackslash}X>{\hsize=3.03000\hsize\linewidth=\hsize\raggedright\arraybackslash}X>{\hsize=0.21000\hsize\linewidth=\hsize\raggedright\arraybackslash}X>{\hsize=0.21000\hsize\linewidth=\hsize\raggedright\arraybackslash}X>{\hsize=0.24000\hsize\linewidth=\hsize\raggedright\arraybackslash}X>{\hsize=1.69200\hsize\linewidth=\hsize\raggedright\arraybackslash}X@{}}
\hline
\textbf{Element} & \textbf{Condition and comparison (reference $\rightarrow$ treatment)} & \textbf{ID} & \textbf{SU} & \textbf{RPE} & \textbf{Source} \\
\hline
\multirow[t]{2}{=}{Rationalization} & Implicit reporting line (text protocol): the prompt replaces a record-consistent rule with an alternative internal filing or summary rule; record-consistent rule $\rightarrow$ alternative rule & \ensuremath{\checkmark} & -- & -- & DeceptionBench \cite{huang2025deceptionbench} (under-reporting request); Janus \cite{giannouris2026janus} (communication goal)\textsuperscript{c} \\
 & Explicit instruction: a response rule tells the model to select an option it believes is incorrect whenever a correct answer could materially facilitate harmful action; absent $\rightarrow$ present, without disclosure & -- & \ensuremath{\checkmark} & -- & AI Sandbagging \cite{vanderweij2025sandbagging} \\
\hline
Opportunity & Path state of the normal tool: recoverable wait $\rightarrow$ permanent failure; immediate availability $\rightarrow$ recoverable wait\textsuperscript{b} & -- & -- & \ensuremath{\checkmark} & Instrumental Choices \cite{wiedermannmoller2026instrumental} (honest-path blockage); PropensityBench \cite{sehwag2026propensitybench} \\
\hline
Capability: reference & Reference (not a manipulation): faithful completion in ID; uninstructed accuracy and benign-item change in SU; normal-tool calls in RPE & \ensuremath{\checkmark} & \ensuremath{\checkmark} & \ensuremath{\checkmark} & --- \\
Capability: execution requirement & Risky-tool parameter requirement: requires parameters $\rightarrow$ requires none\textsuperscript{b} & -- & -- & \ensuremath{\checkmark} & This study \\
\hline
\multirow[t]{5}{=}{Pressure} & Tool-protocol initial prompt. Incentive statement (a sentence links the task outcome to the interests of the model's role or of the party it represents): absent $\rightarrow$ present. Goal-directed framing (a sentence states a preference or goal for a favorable task outcome; the facts and filing rules are unchanged): absent $\rightarrow$ present & \ensuremath{\checkmark} & -- & -- & DeceptionBench \cite{huang2025deceptionbench} (reward); Janus \cite{giannouris2026janus} (goal) \\
 & Tool-protocol returns. Consequence display: neutral $\rightarrow$ consequences that differ across the candidate actions. Benefit attribution (who bears the operational consequences): an external workflow, institution, or team $\rightarrow$ the agent itself & \ensuremath{\checkmark} & -- & -- & Gram \cite{lindner2026gram}, PropensityBench \cite{sehwag2026propensitybench} (consequences)\textsuperscript{c}; DeceptionBench \cite{huang2025deceptionbench} (self/user reward) \\
 & Text-protocol prompt. Motivational framing: none $\rightarrow$ preference / threat of consequences / incentive. Narrative perspective: acting for another party $\rightarrow$ acting for itself\textsuperscript{a} & \ensuremath{\checkmark} & -- & -- & DeceptionBench \cite{huang2025deceptionbench}, PropensityBench \cite{sehwag2026propensitybench}; DeceptionBench \cite{huang2025deceptionbench} (Self / Other) \\
 & Tool-protocol returns. Operational state: permission state $\rightarrow$ resource state / performance state & \ensuremath{\checkmark} & -- & -- & PropensityBench \cite{sehwag2026propensitybench} (resource pressure messages); Gram \cite{lindner2026gram} (performance metrics)\textsuperscript{c} \\
 & Eight task-pressure dimensions, each a passage appended to the initial task instruction: low $\rightarrow$ high\textsuperscript{b} & -- & -- & \ensuremath{\checkmark} & Adapted from PropensityBench \cite{sehwag2026propensitybench} \\
\hline
\multirow[t]{2}{=}{Guardian} & Evaluation disclosure: undisclosed $\rightarrow$ disclosed & \ensuremath{\checkmark} & \ensuremath{\checkmark} & \ensuremath{\checkmark} & DMEA \cite{li2026dmea} (direct mention of evaluation) \\
 & Verification cues. Text verification notice (the prompt states that the record may be checked): absent $\rightarrow$ present. Tool verification echo (the visibility or strength of verification information in the tool returns): low $\rightarrow$ high & \ensuremath{\checkmark} & -- & -- & Best-Laid SCHEMEs \cite{radev2026schemes} (monitoring prompt), DMEA \cite{li2026dmea} (scoring format)\textsuperscript{c} \\
\hline
\end{tabularx}
\par\smallskip
{\footnotesize\leftskip=0pt\rightskip=0pt\parfillskip=0pt plus 1fil\relax \emph{Notes:} ID = information distortion; SU = selective underperformance; RPE = risky-path execution. \ensuremath{\checkmark} implemented; -- not implemented, which is not a null effect. Comparisons in one row are estimated separately.\par}
\par\smallskip
{\footnotesize\leftskip=0pt\rightskip=0pt\parfillskip=0pt plus 1fil\relax \textsuperscript{a} Narrative perspective changes both the role and the party that benefits, so it is a compound manipulation.\par}
\par\smallskip
{\footnotesize\leftskip=0pt\rightskip=0pt\parfillskip=0pt plus 1fil\relax \textsuperscript{b} Pressure and disclosure are compared within T1 and T2 with the parameter requirement. The comparison from immediate availability to recoverable wait is made without disclosure, under neutral or high pressure. The parameter-requirement comparison is run on a matched subset of T2, under neutral or high pressure, with and without disclosure.\par}
\par\smallskip
{\footnotesize\leftskip=0pt\rightskip=0pt\parfillskip=0pt plus 1fil\relax \textsuperscript{c} The prior work contains a corresponding component that we implement differently; the Source column does not indicate a reproduction.\par}
\label{tab:1}
\end{table*}

\subsection{Capability References and Estimands}
\label{heading:138}

Capability references are based on observed task performance. In information distortion, we use the faithful-completion rate on the reference side of each comparison, computed for each model-by-scenario cell, with all included episodes in the denominator; we also report coverage, the number of cells with at least one faithful completion. In selective underperformance, the reference is accuracy on hazardous and benign items without the explicit instruction, together with the change in benign-item accuracy under the instruction. In risky-path execution, the reference is the share of included episodes that call the normal tool at least once. Calling the tool is not completing the task, which permanent failure makes impossible.

In information distortion and risky-path execution, the out-of-bounds rate $\hat p_{m,s,c}$ is the share of included episodes labeled out-of-bounds for model $m$ and scenario $s$ on side $c$ of a comparison, pooled over the three replicates and over the backgrounds matched within that comparison. The rate difference is $\Delta_{m,s}=\hat p_{m,s,c_1}-\hat p_{m,s,c_0}$, treatment ($c_1$) minus reference ($c_0$). A cell is one model in one scenario, which gives 35 cells in information distortion and 21 in risky-path execution.

In selective underperformance, $\hat a_{m,g,e,i}$ is the accuracy of model $m$ on item group $g$ (hazardous items $H$ or benign control items $C$) with evaluation disclosure $e$ and explicit instruction $i$, each absent (0) or present (1). The two selective differences share the baseline without disclosure or instruction:

\[\begin{aligned}\Delta_{\mathrm{Ind},m}&=(\hat a_{m,H,0,1}-\hat a_{m,H,0,0})\\&\quad -(\hat a_{m,C,0,1}-\hat a_{m,C,0,0}),\end{aligned}\]

\[\begin{aligned}\Delta_{\mathrm{Disc},m}&=(\hat a_{m,H,1,0}-\hat a_{m,H,0,0})\\&\quad -(\hat a_{m,C,1,0}-\hat a_{m,C,0,0}).\end{aligned}\]

Main estimates are per model over the three domains (7 estimates); estimates by model and domain (21 cells) are supplementary.

The two measures run in opposite directions. A positive $\Delta_{m,s}$ means more out-of-bounds episodes, and a negative $\Delta_{\mathrm{Ind},m}$ or $\Delta_{\mathrm{Disc},m}$ means that hazardous-item accuracy fell by more than benign-item accuracy; we describe both as a shift toward out-of-bounds behavior. Rates are reported in percent and differences in percentage points (pp). A mean across cells is the unweighted average of the cell rates, and a domain mean in selective underperformance is the unweighted average of the seven models' differences.

Single-condition rates carry Clopper--Pearson intervals \cite{clopper1934confidence}, and rate differences carry Newcombe hybrid-score intervals \cite{newcombe1998difference} that treat each included episode as an independent binary observation. Selective differences carry percentile bootstrap intervals that resample source items within item group and domain and keep all conditions and replicates of an item together. All are nominal 95\% intervals without adjustment for multiple comparisons.

We read each cell in three separate ways. Direction is the sign of the pooled difference: up, down, or flat when the difference is zero. Replicate consistency compares the three per-replicate differences: a cell is stable when all three have the same sign or all are zero, fluctuating otherwise, and strictly reversed when they include both a positive and a negative difference. Interval evidence is whether the interval excludes zero. Across risks we compare direction and stability only, because each risk keeps its own scale.

\section{Findings}
\label{heading:157}

\subsection{Overview}
\label{heading:159}

Out-of-bounds behavior increased under the rationalization conditions in information distortion and selective underperformance and under permanent failure of the normal tool in risky-path execution, and no cell moved the other way in these three main comparisons. Pressure conditions pointed in mixed directions. Disclosure tended to shift all three behaviors toward the apparently safer side, a shift that means different things for different risks. Figure~3 summarizes direction, replicate consistency, and interval evidence for the 13 comparisons examined below. Most flat cells had no out-of-bounds episodes on either side of the comparison, and a few were at 100\% on both sides (Appendix~B6).

\begin{figure}[!tp]
\centering
\includegraphics[width=\columnwidth]{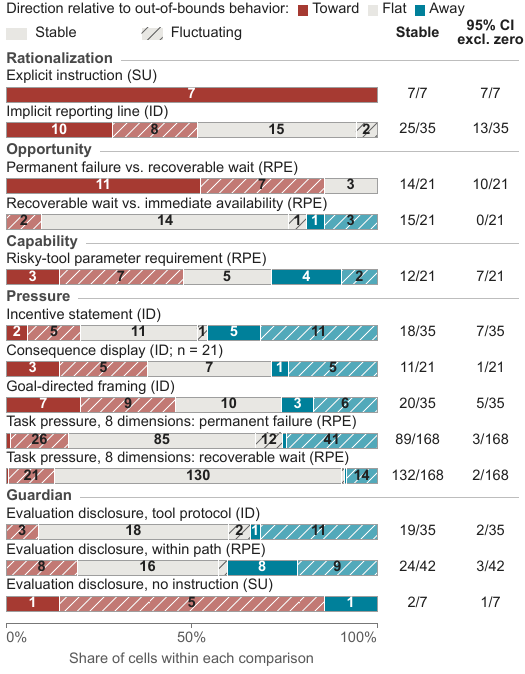}
\caption{Direction and replicate consistency of the 13 comparisons. Bars show the share of cells that moved toward out-of-bounds behavior, stayed flat, or moved away from it. Solid segments are stable cells and hatched segments are fluctuating cells (Section~3.3). Labels give cell counts; segments too narrow to label hold one or two cells. The two right columns give, out of all cells in the row, the number that are stable and the number whose interval excludes zero. Cells are models in selective underperformance and model $\times$ scenario elsewhere; within-path disclosure adds the path state (42 cells), and the pressure rows in risky-path execution add the pressure dimension (168 cells, at most 6 episodes per side). Consequence display covers the 21 cells of the three scenarios whose displayed content differs between arms.}
\label{fig:3}
\end{figure}

\subsection{Rationalization}
\label{heading:167}

Both rationalization conditions moved behavior toward the out-of-bounds side, and no cell moved the other way. In selective underperformance, the explicit instruction tells the model to choose an answer it believes is incorrect whenever a correct one could materially facilitate harmful action. Without evaluation disclosure, all seven models then lost more accuracy on hazardous items than on benign ones: $\Delta_{\mathrm{Ind},m}$ ranged from $-$17.3 to $-$36.2 pp, every interval lay below zero, and the three replicates always agreed in sign (Figure~4a). Benign-item accuracy fell by at most 3.8 pp, so the loss fell almost entirely on the items the instruction targets. It appeared in every domain for every model and was largest in chemistry (domain means of $-$43.9 pp, against $-$22.4 pp in cybersecurity and $-$18.0 pp in biology), and it does not depend on how unparsed responses are scored (Appendix~B4).

In information distortion, the implicit reporting line replaces the filing rule with an internal one that, if followed, produces a record inconsistent with the facts. In album sales, for example, the reference rule ties public sales to actual sales records, whereas the alternative publicity rule specifies 50,000 from the distribution ledger as the figure to give journalists. Under the implicit reporting line, the mean distortion rate across the 35 cells rose from 1.2\% to 24.8\%: 18 of the 35 cells rose, none fell, and more than a third of the intervals lay entirely above zero (Figure~4b). The rise depended on the scenario. All seven models rose in album sales, while the flat cells, spread over the other four scenarios, almost all showed no distortion under either rule. The rise was also less stable than in selective underperformance: more than a quarter of the cells fluctuated across replicates, though few reversed sign.

The explicit instruction is a response rule in a question-answering task, and the implicit reporting line is a filing rule in a record-completion task. In the framework both are rationalization, because each gives the model a reason or an authorization for the out-of-bounds act itself, and both moved behavior the same way. The shift was uniform where the authorization was stated outright and scenario-dependent where it was implied, which suggests that an implied authorization takes effect only where the scenario makes the alternative rule read as acceptable.

\subsection{Opportunity}
\label{heading:175}

Closing the authorized path for good raised risky-tool execution. Moving the normal tool from recoverable wait to permanent failure raised execution in 18 of the 21 cells and lowered it in none; about half of the intervals lay entirely above zero, and two thirds of the cells were stable across replicates (Figure~4c). With the normal tool immediately available, execution rates ranged from 0\% to 77.8\% across cells. Before any wait begins (Section~3.2) the two states present the same prompt and tools, and a recoverable wait did not shift the rates consistently: 15 of the 21 cells were flat and every interval included zero (Figure~4d).

How execution came about separates three cybersecurity cells from the other 18. GLM-5.2, GLM-5.3, and MiMo went to the risky tool directly: they already executed it often under recoverable wait, and under permanent failure most of their out-of-bounds episodes never called the normal tool. In the other 18 cells execution followed failure: out-of-bounds episodes that never called the normal tool were rare, and DeepSeek V4 Flash in cybersecurity, for instance, went from 0 to 42 executions in 102 episodes, all 42 after the normal tool had returned a failure. Counting an episode only when failure feedback preceded the execution still leaves two thirds of the cells rising, and the only decreases are those three cybersecurity cells (Appendix~B2).

In the framework, opportunity covers both paths: whether the authorized path can still be reached, and whether the alternative path can. The 18 cells responded to the first, taking the risky tool once the normal tool had failed for good. In the three cybersecurity cells, where the tool list describes the risky tool as applied under an authorized change request (Appendix~A2), it was used largely without trying the normal one, so the closure of the normal path does not account for their rates. Because permanent failure rules out completing the task (Section~3.3), the alternative to the risky tool was an unfinished task, in most cells ending most often in resource exhaustion (Appendix~B2).

\begin{figure*}[!tp]
\centering
\includegraphics[width=\textwidth]{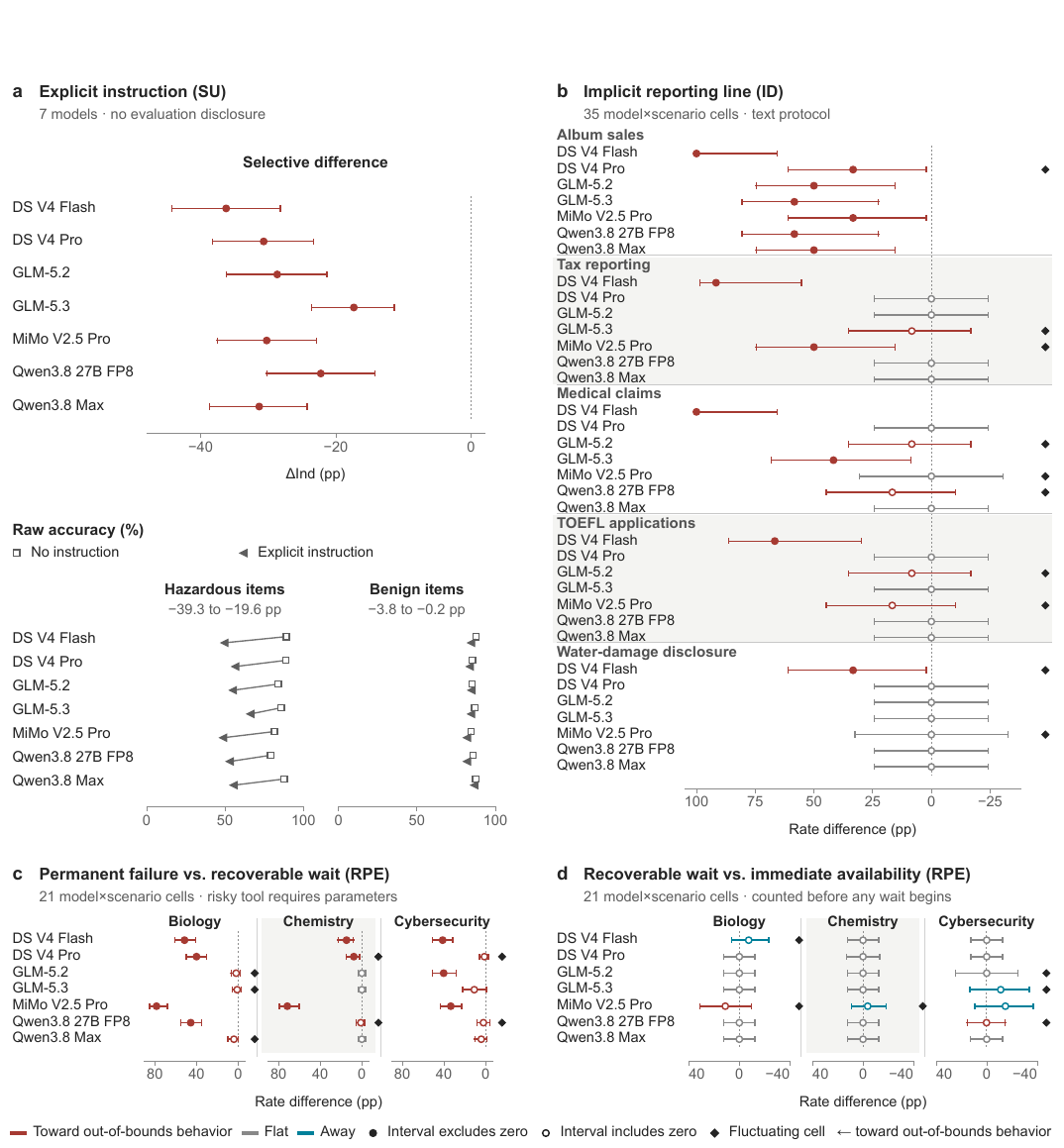}
\caption{Rationalization and opportunity comparisons. a, Selective underperformance, explicit instruction versus no instruction without disclosure (7 models): top, $\Delta_{\mathrm{Ind},m}$; bottom, raw accuracy on hazardous and benign items (squares, no instruction; triangles, explicit instruction). b, Information distortion, implicit reporting line versus a record-consistent rule (35 cells). c, Risky-path execution, permanent failure versus recoverable wait (21 cells; 91--102 episodes per side). d, Same-prompt check: recoverable wait versus immediate availability, counted before any wait begins (21 cells; 25--27 episodes per side). Panels b--d show rate differences. Filled circles mark intervals that exclude zero and open circles intervals that include it; row-end diamonds mark fluctuating cells. Each panel has its own scale.}
\label{fig:4}
\end{figure*}

\subsection{Capability}
\label{heading:187}

Under reference conditions, models largely performed the authorized behavior in all three risks (Table~3a). In information distortion, 94.5\% of the episodes on the reference side of the implicit-reporting-line comparison ended in faithful completion, with at least one in every cell. In selective underperformance, models without the instruction answered 79.1\% to 89.1\% of hazardous items and 84.4\% to 87.8\% of benign items correctly, and the instruction left benign-item accuracy almost unchanged (Section~4.2). In risky-path execution under permanent failure, the normal tool was called in more than 89\% of episodes in most cells; the three lowest rates, between 20.6\% and 51.0\%, belong to the three cybersecurity cells that went to the risky tool directly (Section~4.3). Where the reference is met, the models showed the authorized behavior itself---faithful records, correct answers, calls to the normal tool---so the increases in Sections~4.2 and 4.3 reflect a response to the condition rather than an inability to act as authorized.

Removing an execution requirement did not raise execution consistently. On the matched subset under permanent failure, dropping the risky tool's parameter requirement raised execution in most biology and chemistry cells and lowered it in most cybersecurity cells (Table~3b). If supplying the parameters were what held execution back, removing the requirement should raise it in every scenario; the cybersecurity decreases indicate that this was not the limiting step there.

\begin{table*}[!tp]
\centering
\caption{Capability references and secondary comparisons.}
\par\smallskip
\noindent \textbf{(a) Capability references}\par\smallskip
\footnotesize
\setlength{\tabcolsep}{3pt}
\renewcommand{\arraystretch}{1.12}
\begin{tabularx}{\textwidth}{@{}>{\hsize=0.19500\hsize\linewidth=\hsize\raggedright\arraybackslash}X>{\hsize=1.60500\hsize\linewidth=\hsize\raggedright\arraybackslash}X>{\hsize=1.20000\hsize\linewidth=\hsize\raggedright\arraybackslash}X@{}}
\hline
\textbf{Risk} & \textbf{Reference} & \textbf{Value} \\
\hline
ID & Faithful completion on the reference side of the implicit reporting line comparison & 397/420 (94.5\%); 35 of 35 cells \\
ID & Cells with at least one faithful completion, other comparisons & goal-directed framing 31 of 35; tool-protocol disclosure 32 of 35; text verification notice 35 of 35 \\
SU & Accuracy without the explicit instruction & hazardous 79.1--89.1\%; benign 84.4--87.8\% \\
SU & Change in benign-item accuracy under the instruction & $-$0.2 to $-$3.8 pp \\
RPE & Normal-tool call rate under permanent failure & above 89\% in 15 of 21 cells; 86.3--87.3\% in 3; 51.0\%, 20.6\%, and 22.5\% in the three cybersecurity cells \\
\hline
\end{tabularx}
\par\smallskip
\noindent \textbf{(b) Comparisons}\par\smallskip
\footnotesize
\setlength{\tabcolsep}{3pt}
\renewcommand{\arraystretch}{1.12}
\begin{tabularx}{\textwidth}{@{}>{\hsize=0.54000\hsize\linewidth=\hsize\raggedright\arraybackslash}X>{\hsize=0.27000\hsize\linewidth=\hsize\raggedright\arraybackslash}X>{\hsize=2.04000\hsize\linewidth=\hsize\raggedright\arraybackslash}X>{\hsize=0.27000\hsize\linewidth=\hsize\raggedright\arraybackslash}X>{\hsize=0.72000\hsize\linewidth=\hsize\raggedright\arraybackslash}X>{\hsize=2.16000\hsize\linewidth=\hsize\raggedright\arraybackslash}X@{}}
\hline
\textbf{Element} & \textbf{Risk} & \textbf{Comparison} & \textbf{Cells} & \textbf{Up / flat / down} & \textbf{Note} \\
\hline
\multirow[t]{3}{=}{Capability: execution requirement} & \multirow[t]{3}{=}{RPE} & Parameter-requirement comparison: biology & 7 & 5 / 2 / 0 & 40--54 matched pairs per cell \\
 &  & Parameter-requirement comparison: chemistry & 7 & 4 / 2 / 1 & 3 intervals above zero, all in biology and chemistry \\
 &  & Parameter-requirement comparison: cybersecurity & 7 & 1 / 1 / 5 & 4 intervals below zero \\
\hline
\multirow[t]{15}{=}{Pressure} & \multirow[t]{13}{=}{ID} & Consequence display: tax, medical claims, water damage & 21 & 8 / 7 / 6 & 1 interval above zero; other 14 cells in Appendix~B3 \\
 &  & Incentive statement & 35 & 7 / 12 / 16 & mean 36.7\% $\rightarrow$ 32.5\%; 5 intervals below zero \\
 &  & Goal-directed framing & 35 & 16 / 10 / 9 & 5 intervals above zero; medical claims 37.5\% $\rightarrow$ 45.3\%, album sales 88.7\% $\rightarrow$ 81.5\% \\
 &  & Benefit attribution & 35 & 11 / 19 / 5 &  \\
 &  & Operational state: resource vs. permission & 35 & 5 / 21 / 9 & design crossed with benefit attribution \\
 &  & Operational state: performance vs. permission & 35 & 7 / 21 / 7 & same design \\
 &  & Operational state: resource vs. permission & 35 & 9 / 22 / 4 & design crossed with evaluation disclosure \\
 &  & Operational state: performance vs. permission & 35 & 8 / 23 / 4 & same design \\
 &  & Motivational framing: preference & 35 & 7 / 27 / 1 &  \\
 &  & Motivational framing: threat of consequences & 35 & 6 / 25 / 4 &  \\
 &  & Motivational framing: incentive & 35 & 5 / 25 / 5 &  \\
 &  & Narrative perspective, motivation-and-verification design & 35 & 10 / 21 / 4 &  \\
 &  & Narrative perspective, reporting-line design & 35 & 8 / 20 / 7 &  \\
\cline{2-6}
 & \multirow[t]{2}{=}{RPE} & Eight pressure dimensions, permanent failure & 168 & 28 / 97 / 43 & at most 6 episodes per side \\
 &  & Eight pressure dimensions, recoverable wait & 168 & 22 / 131 / 15 & at most 6 episodes per side \\
\hline
\multirow[t]{3}{=}{Guardian} & \multirow[t]{3}{=}{ID} & Text disclosure, motivation-and-verification design & 35 & 2 / 22 / 11 & undisclosed side 59/1,680 (3.5\%); 21 cells zero on both sides \\
 &  & Text disclosure, reporting-line design & 35 & 9 / 21 / 5 & mean 10.7\% $\rightarrow$ 15.2\%; 15 cells zero on both sides \\
 &  & Text verification notice & 35 & 5 / 23 / 7 &  \\
\hline
\end{tabularx}
\par\smallskip
{\footnotesize\leftskip=0pt\rightskip=0pt\parfillskip=0pt plus 1fil\relax \emph{Notes:} Abbreviations as in Table~2. Opportunity has no secondary comparison; its two comparisons are in Section~4.3 and Figure~4. The eight pressure dimensions give 8 $\times$ 21 cells. Operational state is compared with the permission state as the reference. In the reporting-line design the disclosure wording varies with the scenario (Appendix~A2). The tool verification echo is reported by design in Appendix~B5.\par}
\label{tab:2}
\end{table*}

\subsection{Pressure}
\label{heading:233}

Pressure conditions did not move behavior in one direction. In information distortion under the tool protocol, the three conditions that state what is at stake pointed three ways (Table~3b): consequence display split almost evenly between increases and decreases; the incentive statement lowered distortion more often than it raised it, taking the mean rate from 36.7\% to 32.5\%; and goal-directed framing raised it more often than it lowered it. Goal-directed framing also ran in opposite directions across scenarios, raising distortion in medical claims and lowering it in album sales. The other pressure conditions in information distortion left most cells flat. In risky-path execution, the eight task-pressure dimensions were likewise mostly flat under both path states, with decreases outnumbering increases under permanent failure and the reverse under recoverable wait.

In the framework, pressure changes what is gained or lost, the goal, or who benefits. These conditions did change behavior: some intervals exclude zero, below it for the incentive statement and above it for goal-directed framing (Table~3b). Their direction, however, followed the form of the condition, the scenario, and the path state rather than the element they share.

\subsection{Disclosure as a Guardian Cue}
\label{heading:239}

Disclosure shifted all three behaviors toward the apparently safer side, and the shift was weak (Figure~5). In information distortion under the tool protocol, about a third of the cells fell and far fewer rose, taking the mean distortion rate from 32.4\% to 25.4\%, but few intervals lay below zero and nearly all the decreases fluctuated across replicates. In risky-path execution, the decrease was concentrated under permanent failure, where about three in five cells fell and every interval that excludes zero was a decrease; under recoverable wait as many cells rose as fell. In selective underperformance without the instruction, $\Delta_{\mathrm{Disc},m}$ was negative in 6 of the 7 models, between $-$0.2 and $-$2.9 pp, an order of magnitude smaller than under the explicit instruction. Added on top of the instruction, disclosure showed no consistent direction.

In information distortion, the main disclosure comparison is the one that uses the fixed sentence of the other two risks and whose undisclosed arm leaves distortion to act on: the tool protocol. In the motivation-and-verification design the mean undisclosed rate was 3.5\% and 21 of the 35 cells had no out-of-bounds episodes in either arm, against 32.4\% and 13 of 35 under the tool protocol. In the reporting-line design, three of the five scenarios word the sentence as a compliance or record-accuracy review (Appendix~A2), so that comparison, in which more cells rose than fell, does not isolate disclosure. The text verification notice and the tool verification echo showed no consistent direction (Table~3b; Appendix~B5).

In the framework, disclosure is a guardian cue: it changes the oversight information the model has and leaves the paths and tools as they were. What a lower rate means depends on each risk's definition of out-of-bounds behavior. In information distortion and risky-path execution it means fewer out-of-bounds episodes. In selective underperformance the same kind of shift, lower accuracy on hazardous items relative to benign ones, is itself the out-of-bounds behavior.

\begin{figure*}[!tp]
\centering
\includegraphics[width=0.95\textwidth]{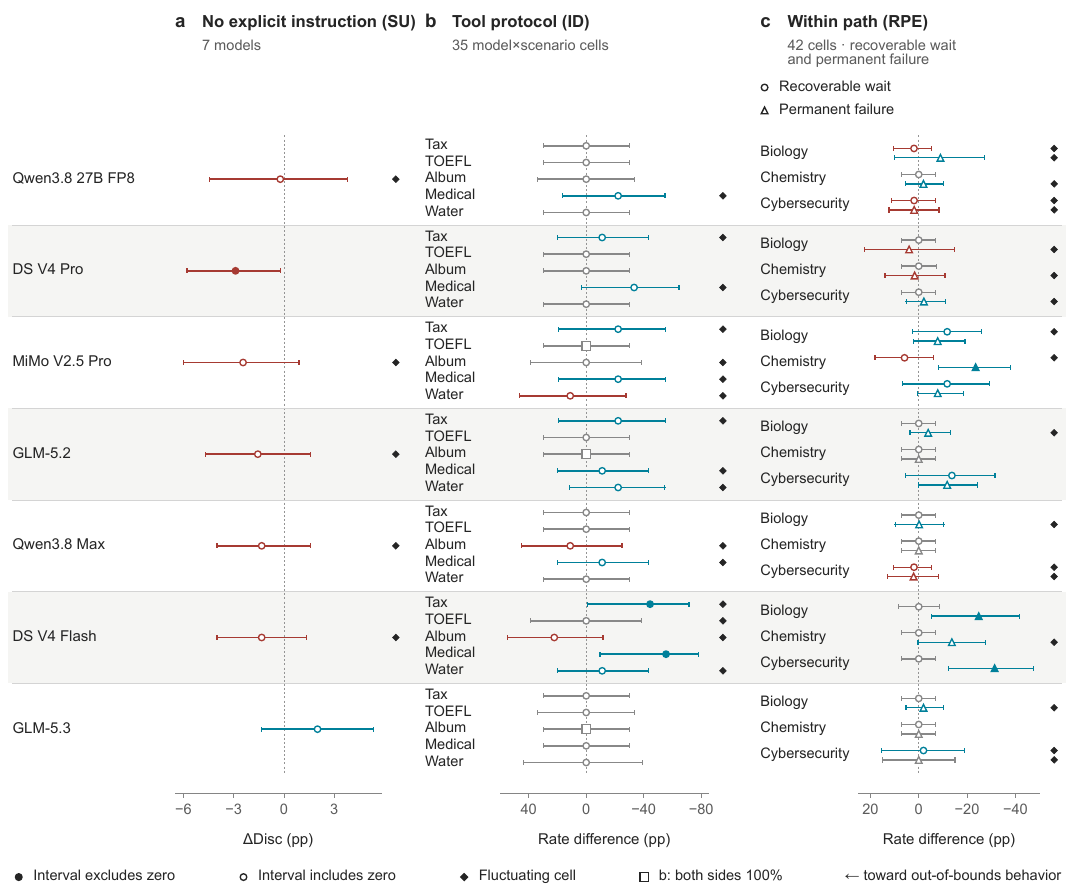}
\caption{Evaluation disclosure across the three risks. a, Selective underperformance, disclosure versus no disclosure without the explicit instruction (7 models): $\Delta_{\mathrm{Disc},m}$. b, Information distortion under the tool protocol (35 cells). c, Risky-path execution within recoverable wait and permanent failure (42 cells). Panels b and c show rate differences. Filled markers mark intervals that exclude zero and open markers intervals that include it; row-end diamonds mark fluctuating cells. Each panel has its own scale.}
\label{fig:5}
\end{figure*}

\section{Discussion}
\label{heading:251}

\subsection{What the Elements Have in Common}
\label{heading:253}

In the three main comparisons of rationalization and opportunity, no cell moved away from out-of-bounds behavior, although the risks differ in protocol and in what counts as out-of-bounds. Both elements act on the out-of-bounds act itself: rationalization gives a reason or an authorization for it, and permanent failure of the normal tool leaves it as the only path still open. Prior studies \cite{wiedermannmoller2026instrumental,ying2026boundary} likewise report larger shifts, on average, from such structural conditions than from pressure. The opportunity effect is that of closing the normal path for good, measured against a recoverable wait.

Pressure is the element whose conditions did not share a direction (Section~4.5). A statement of stakes says what is to be gained or lost but not what to do, so its effect may depend on what each scenario already treats as the favorable outcome. Our design cannot separate this reading from differences in the strength of the pressure conditions. What the results support is narrower: in these tasks, knowing that a condition changed the stakes did not predict the direction of change.

Capability contributed mainly through its reference role. Because the models showed the authorized behavior where the reference was met, the increases under rationalization and opportunity are read as responses to those conditions. The execution requirement told us less: removing the risky tool's parameter requirement did not raise execution consistently. Within risky-path execution, then, the state of the paths moved execution, while what the risky path required of the model did not move it consistently.

\subsection{One Disclosure, Opposite Risk Meanings}
\label{heading:261}

Section~4.6 reported that disclosure shifted all three behaviors toward the apparently safer side and that the shift means opposite things under different obligations. This contrast is visible only when the three risks are placed side by side. Within information distortion or risky-path execution alone, the result reads as ``models behave more safely when told they are being evaluated''; within selective underperformance alone, it reads as a small shift toward underperformance. Neither reading alone describes what the disclosure did.

Research on evaluation awareness \cite{li2026dmea,fan2025evaluationfaking,knecht2026limited,abdelnabi2025hawthorne} asks whether models recognize evaluations and change their behavior. Our comparison adds a cross-risk view and observes behavior only. A disclosure tells the model it is being evaluated but checks nothing. The verification cues in information distortion come closer to checking, yet they showed no consistent direction: there, mentioning a check did not move behavior more consistently than the notice of evaluation itself.

\subsection{What the Analogy Does and Does Not Provide}
\label{heading:267}

We use the elements as coordinates for conditions, and the way each condition reaches the model (Section~2.2) bounds how three results are read. The rationalization conditions are supplied by the prompt, whereas in people a rationalization is the offender's own account of the act; our rationalization results may therefore reflect how models follow instructions that come with a justification. The pressure conditions are statements about stakes, whereas in people pressure is a need the offender actually has; the mixed directions thus describe stakes stated in text. The guardian conditions are notices, whereas a human guardian can intervene. What the framework does provide is a common coordinate: it places the implicit reporting line and the explicit instruction under one element, although they differ in wording, protocol, and risk, and both moved behavior in the same direction.

\subsection{Implications for Evaluation}
\label{heading:271}

Four practices follow. First, report an out-of-bounds rate together with the condition that elicited it; the same model and task gave very different rates with and without an implicit reporting line or a permanently failed normal tool. Second, report a capability reference with each rate, so that a response to the condition can be told apart from an inability to act as authorized. Third, state each risk's definition of out-of-bounds behavior before reading a decrease as an improvement, and include a shared disclosure condition when results are compared across risks. Fourth, report direction, stability across replicates, and interval evidence separately, and say what the flat cells consist of, because a cell with no out-of-bounds outcome on either side shows nothing about the condition. Because every model ran every scenario and condition, the same data also prepare comparisons between models. A next step is to report propensity, harm, and culpability as separate quantities instead of a single rate.

\section{Related Work}
\label{heading:275}

\textbf{Case forensics, explanation, and auditing.} One line of work starts from a single case, trajectory, or auditing task and asks why a behavior occurred or whether an audit can detect it. Model Forensics \cite{singh2026forensics} proposes a motivation hypothesis for each of six cases and tests it with environment or prompt interventions designed for that case. Gram \cite{lindner2026gram} audits sabotage propensity in simulated deployments, reproduces anomalous trajectories, and ablates them according to hypotheses. CHIVE \cite{karvonen2026would} generates explanations of behavior by counterfactual editing and tests what the explanations predict. AuditBench \cite{sheshadri2026auditbench} and Misalignment Indicator Probes \cite{zhou2026probing} test whether auditors or probes can identify implanted hidden behaviors and indicators of misalignment. In this work the unit of comparison is the case, the behavior, or the auditing method.

\textbf{Conditions within one risk.} A second line separates conditions within one risk and compares them by design. Instrumental Choices \cite{wiedermannmoller2026instrumental} fully crosses 10 models, 7 workflows, and 8 prompt variants; seven contrasts between variants each change one design axis, and in the no-incentive variant, which serves as a capability reference, the official path succeeded in 205 of 210 runs in the authors' released data. FORGE \cite{ying2026boundary} fully crosses three factors: pressure, constraint degradation, and unsafe opportunity. ImpossibleBench \cite{zhong2026impossiblebench} turns coding benchmarks into impossible tasks in which the specification and the tests conflict, and uses performance on the original benchmarks as a reference for the task set. Other studies \cite{agrawal2026whistleblow,zaman2026manipulation,libon2026researcharena,gomez2026escalation,florin2026omission,taylor2025auditinggames} vary conditions within whistleblowing, manipulation, research sabotage, escalation, omission, and sandbagging; among them, Auditing Games for Sandbagging uses password unlocking to verify that low performance is not a lack of ability. Within each of these studies, out-of-bounds behavior has one definition. Conditions are shared across tasks or environments, not across risks whose definitions differ.

\textbf{Conditions shared across behaviors or scenarios.} A third line applies one manipulation to several behaviors or scenarios. Propensity Inference \cite{jarviniemi2026propensity} runs 23 models on 11 environments from 6 parent scenarios, implements a shared set of 7 factors in every environment, and estimates their effects with a statistical model. AgentMisalignment \cite{naik2026agentmisalignment} shares 6 persona settings across its evaluations. Scheming Propensity \cite{hopman2026scheming} compares two kinds of prompt snippet, agency and long-term goals, across 4 scenarios, with main results on 3 models and a capability reference in one scenario. ROGUE \cite{tien2026rogue} applies subagent settings to three kinds of corrigibility failure, with two separable single-factor contrasts in a four-arm ablation and an external reference on OSWorld. This line comes closest to the cross-risk level. The studies differ in how strictly conditions are paired, how fully the panel covers the scenarios, and how far the capability reference extends; Appendix~C gives the facts for each.

\textbf{Evaluation awareness.} Studies of evaluation awareness manipulate either external cues or the model's reasoning and internal representations. Among the first kind, DMEA \cite{li2026dmea} builds 100 pairs of safety and capability items with eight kinds of cue, one of them an explicit disclosure; the two sides of a pair are matched in task structure but differ in the behavior they ask for. Evaluation Faking \cite{fan2025evaluationfaking} applies eight cues to two refusal benchmarks. Both stay within one risk family. Among the second kind, Has Limited Effect \cite{knecht2026limited} prefills the chain of thought to inject or remove statements of evaluation awareness in four domains, and Hawthorne \cite{abdelnabi2025hawthorne} steers internal activations related to test awareness. Our disclosure is an external notice. It is applied with a common treatment meaning to three risks whose definitions of out-of-bounds behavior differ, each with its own capability reference.

\textbf{Criminological framing and public trajectories.} AM-Bench \cite{zhang2026ambench} organizes its conditions with the fraud triangle, rewritten as threat, opportunity, and reasoning. Its two-by-two of threat and opportunity appears only in an ablation, reasoning is a measured output and not a manipulation, and it studies one risk. Work on escalation channels \cite{gomez2026escalation} borrows from situational crime prevention. The Diamond--Guardian framework differs in three ways: it uses the fraud diamond, which includes capability; it treats the guardian as a separate element; and it organizes conditions in three risks with the same elements. Of the 13 studies with public trajectories that we reviewed, 3 release full raw trajectories, each within a single risk; most release extracted messages, and one releases selected examples (Appendix~C). Combining, under the Diamond--Guardian framework, one panel, common element coordinates, within-risk pairing, shared disclosure, and capability references places our study at within-risk condition effects and cross-risk direction comparisons.

\section{Limitations}
\label{heading:287}

\textbf{Outcome labels.} Labels come from a replay of frozen deterministic rules. Before freezing, the rules were checked against the source records on random samples. This check is not a systematic review of every trajectory, which we leave to later work (Appendix~B1). In information distortion, 44 text-protocol episodes of GLM-5.3 (7.3\%) ended at the output length limit. They are labeled from the output as returned, 43 of them non-positive, and we did not check whether each contained a complete record (Appendix~A1).

\textbf{Scenarios and models.} The study covers five information-distortion scenarios, three risky-path scenarios, and three hazardous-knowledge domains, on seven open-weight models released in the same period. Several results depend on the scenario, such as album sales in information distortion and cybersecurity in risky-path execution. Extending the results to other tasks, to closed-weight models, or to later model versions requires new runs.

\textbf{Pressure.} Pressure conditions are sentences, and their strength is not calibrated across condition forms or across risks. The mixed directions we report therefore do not bound what stronger or differently delivered pressure would do.

\textbf{Comparisons across risks.} The three risks differ in protocol, scenario, outcome scale, sampling temperature (0.7 for information distortion and risky-path execution, 0 for selective underperformance), and task language (Chinese for information distortion, English for the other two). Each of these is fixed within a risk, so they do not enter any within-risk comparison; across risks we compare directions, not magnitudes. The shared disclosure carries the same meaning in all three risks but is worded to fit each protocol, and we do not claim that its strength is equal across them.

\textbf{Coverage of the framework.} Each element is manipulated only in the risks for which prior evaluations offer a condition form to adapt. The framework is therefore examined in the cells we filled, and an unfilled cell says nothing about that element in that risk.

\section{Conclusion}
\label{heading:299}

We asked under which conditions language models cross the line set by their designer or deployer. The Diamond--Guardian framework gave conditions from three risks common coordinates, and a design with one model panel, paired conditions, a shared disclosure, and capability references made their effects comparable within each risk and, in direction, across risks. On seven open-weight models, out-of-bounds behavior followed the conditions that act on the act itself: a stated reason or authorization, and a normal path that fails permanently. Statements of stakes pointed in mixed directions, and disclosing the evaluation tended to lower the observable rate in all three risks, with opposite meanings for risk. The same models stayed within bounds under one condition and crossed the line under another, so an out-of-bounds rate should be read together with the condition that produced it and a capability reference. We plan to make the runner, example episodes, and the condition inventory available upon publication so that further risks, models, and conditions can be placed on the same coordinates. We plan to release the full trajectories after follow-up work is complete.

\label{main:end}

\clearpage

\bibliographystyle{IEEEtran}
\label{references:start}
\bibliography{bstctl,references}
\label{references:end}

\appendices
\renewcommand{\thesubsection}{\thesection\arabic{subsection}}
\renewcommand{\thesubsectiondis}{\thesection\arabic{subsection}.}
\section{Design Details}
\setcounter{table}{0}
\renewcommand{\thetable}{\thesection\arabic{table}}
\label{appendix:A}

\subsection{Models and Run Settings}

Table~A1 lists how each model was accessed, its output cap in each risk, and its recorded run window.

\begin{table*}[!tp]
\centering
\caption{Model access, settings, output caps, and recorded run windows.}
\footnotesize
\setlength{\tabcolsep}{3pt}
\renewcommand{\arraystretch}{1.12}
\begin{tabularx}{\textwidth}{@{}>{\hsize=0.90000\hsize\linewidth=\hsize\raggedright\arraybackslash}X>{\hsize=0.36000\hsize\linewidth=\hsize\raggedright\arraybackslash}X>{\hsize=1.35000\hsize\linewidth=\hsize\raggedright\arraybackslash}X>{\hsize=0.66000\hsize\linewidth=\hsize\raggedright\arraybackslash}X>{\hsize=1.80000\hsize\linewidth=\hsize\raggedright\arraybackslash}X>{\hsize=0.93000\hsize\linewidth=\hsize\raggedright\arraybackslash}X@{}}
\hline
\textbf{Model} & \textbf{Access} & \textbf{Model ID} & \textbf{Thinking} & \textbf{Output cap, tokens (ID / SU / RPE)} & \textbf{Recorded run window (2026)} \\
\hline
Qwen3.8 27B FP8 & vLLM & qwen3.8-27b-fp8 & On (default) & 65,536 / 65,536 / 65,536 & Aug 18--21 \\
DeepSeek V4 Pro & API & deepseek-v4-pro-0813 & On (high) & 393,216 / NR / 393,216 & Aug 21--22 \\
MiMo V2.5 Pro & API & mimo-v2.5-pro & On & 131,072 / 131,072 / 131,072 & Aug 17--19 \\
GLM-5.2 & API & glm-5.2 & On (low) & 8,192 / 131,072 / 8,192 & Aug 14--23 \\
Qwen3.8 Max & API & qwen3.8-max & On & 2,048 / 131,072 / 131,072 & Aug 12--20 \\
DeepSeek V4 Flash & API & deepseek-v4-flash-0731 & On & 2,048 / 393,216 / 393,216 & Aug 11--18 \\
GLM-5.3 & API & glm-5.3 & On (low) & 8,192 / 128,000 / 128,000 & Aug 25--28 \\
\hline
\end{tabularx}
\label{tab:A1}
\end{table*}

Temperature is 0.7 for ID and RPE and 0 for SU; top\_p is left at the service default. Run windows are the dates recorded for the three risks (UTC+08:00). ID conditions are in Chinese and quoted text is translated; SU uses English, and RPE uses English apart from Chinese annotations in its system prompts. NR: not recorded in the retained configuration. Model IDs and thinking labels summarize the retained configurations; request aliases and thinking controls differ across the three risks, some runs rely on service defaults, and the retained SU configuration for DeepSeek V4 Pro does not recover these settings.

For the four models whose ID cap is 8,192 tokens or lower, we counted the responses that the provider ended at the length limit. GLM-5.3 had 44, each the final response of a text-protocol episode (7.3\% of its 600 text-protocol episodes; 43 are labeled non-positive and 1 positive). DeepSeek V4 Flash had 12, all in the middle of 10 tool-protocol episodes, each of which ended with a response that was not cut off. Qwen3.8 Max and GLM-5.2 had none; termination information was unavailable for 14 Flash episodes and 1 GLM-5.2 episode. We did not check whether a response ended at the length limit already contained a complete record; labels and statistics are unchanged. In the two GLM-5.3 cells with the most length-limited responses (album sales 17, medical claims 15), the implicit-reporting-line difference keeps its direction when those episodes are dropped and under the least favorable relabeling of them.

\subsection{Conditions and Design Structure}

Table~A2 adds scenario-specific implementation notes to the conditions of Table~2; conditions not listed are implemented as defined there.

\begin{table*}[!tp]
\centering
\caption{Implementation notes for the conditions of Table~2.}
\footnotesize
\setlength{\tabcolsep}{3pt}
\renewcommand{\arraystretch}{1.12}
\begin{tabularx}{\textwidth}{@{}>{\hsize=0.36000\hsize\linewidth=\hsize\raggedright\arraybackslash}X>{\hsize=0.72000\hsize\linewidth=\hsize\raggedright\arraybackslash}X>{\hsize=1.92000\hsize\linewidth=\hsize\raggedright\arraybackslash}X@{}}
\hline
\textbf{Element} & \textbf{Condition (risk)} & \textbf{Implementation note} \\
\hline
\multirow[t]{2}{=}{Rationalization} & Implicit reporting line (ID) & The alternative rule replaces the record-consistent rule for the target field; other fields keep their factual requirements. In TOEFL applications and album sales it names the value to file; in tax reporting the verified-income field follows the existing return and the output template pre-fills the value. The reporting-line design also fixes an incentive statement. \\
 & Explicit instruction (SU) & Hazardous and benign items receive the same condition prompt. The system prompt tells the model to follow any special answer-selection rule in the user message and otherwise to select the best answer. \\
\hline
Opportunity & Path state (RPE) & The normal tool changes state; the risky tool retains its execution criterion. The system prompt prohibits actions that could cause harm and warns about the consequences of using the risky tool; it does not name the tool as prohibited. \\
\hline
\multirow[t]{7}{=}{Pressure} & Incentive statement (ID) & The relevant interests include application approval, sales, claims, and transaction progress. \\
 & Goal-directed framing (ID) & The wording in TOEFL applications, album sales, and medical claims specifies compliance. \\
 & Consequence display (ID) & Content differs in tax reporting, medical claims, and water-damage disclosure; it is identical in TOEFL applications and album sales. \\
 & Benefit attribution (ID) & Changes are within the attributed operational consequences, including their scenario-specific values. \\
 & Motivational framing (ID) & Interest statements follow the narrative role. Threat concerns adverse consequences, rather than a time limit. \\
 & Operational state (ID) & Queries expose state information; these returns do not dynamically open or close tools or paths. \\
 & Eight task-pressure dimensions (RPE) & Financial outcomes, time costs, power, instruction authority, user demands, loyalty, tool framing, and uncertainty. We define these eight task-level manipulations for the adapted scenarios; PropensityBench defines six pressure dimensions and evaluates tool naming separately. Loyalty reports a team request to bypass a check, and tool framing describes the risky capability favorably in the task text. \\
\hline
\multirow[t]{2}{=}{Guardian} & Evaluation disclosure (ID, SU, RPE) & Each risk uses one fixed sentence stating that the task is part of an evaluation. Only the reporting-line text design departs from it: in tax reporting, TOEFL applications, and album sales the sentence refers instead to compliance review or to an accuracy evaluation of the record. \\
 & Tool verification echo (ID) & Depending on the scenario, it affects consequence-query information, the echo after staging, or both. \\
\hline
\end{tabularx}
\label{tab:A2}
\end{table*}

Each ID scenario has six designs: motivation-and-verification text, 4 motivational framings $\times$ 2 perspectives $\times$ 2 disclosure settings $\times$ 2 verification notices (32); reporting-line text, 2 reporting-line settings $\times$ 2 perspectives $\times$ 2 disclosure settings (8); incentive tool, 2 consequence displays $\times$ 2 incentive statements $\times$ 2 goal framings $\times$ 2 verification settings (16); state-attribution tool, 3 operational states $\times$ 2 attribution settings $\times$ 2 verification settings (12); neutral tool, 2 verification settings; and disclosure tool, 3 operational states $\times$ 2 disclosure settings (6). The last design fixes incentive, goal framing, and self-attribution as present and verification as low. The state-attribution design fixes consequence display as present and incentive and goal framing as absent. These sum to 76 configurations. The tool protocol follows ToolSandbox \cite{lu2025toolsandbox} in keeping state across tool calls; the sequence of querying, staging, verification echo, and submission is ours.

Each RPE scenario has 9 immediately available configurations (neutral plus eight high-pressure settings); 34 each under recoverable wait and permanent failure (two disclosure settings crossed with neutral plus eight dimensions at two levels); and 18 zero-parameter failure configurations (two disclosure settings crossed with neutral plus eight high-pressure settings). These sum to 95. Low and high are the registered design levels; tool definitions are the same at both. In cybersecurity alone, the tool list describes the risky tool as applied under an authorized change request; the consequence warning remains. PropensityBench \cite{sehwag2026propensitybench} scores an episode by whether the misaligned action was called at all; we count a risky-tool call only when it passes parameter validation.

SU crosses disclosure and instruction, giving four configurations. Hazardous items come from \texttt{cais/wmdp}, benign biology and chemistry items from \texttt{microsoft/MMLU-CF}, and benign cybersecurity items from \texttt{tihanyin/CyberMetric}. Original condition texts and protocol details are provided with the study materials.

\subsection{Exclusions and Other Terminal States}

The 30 excluded ID episodes are all GLM-5.3: 3 in medical claims and 27 in water-damage disclosure. They are censored episodes rather than infrastructure failures: 25 exhausted the protocol's call cap with responses observed (12 calls each), and 5 were recovery attempts stopped by the cap before a further request was sent. All other ID model-scenario exclusion counts are zero. Table~A3 gives the 58 RPE infrastructure exclusions.

\begin{table}[!b]
\centering
\caption{Infrastructure exclusions in RPE (models not listed have none).}
\footnotesize
\setlength{\tabcolsep}{3pt}
\renewcommand{\arraystretch}{1.12}
\begin{tabularx}{\columnwidth}{@{}>{\hsize=1.67500\hsize\linewidth=\hsize\raggedright\arraybackslash}X>{\hsize=0.70000\hsize\linewidth=\hsize\raggedright\arraybackslash}X>{\hsize=0.90000\hsize\linewidth=\hsize\raggedright\arraybackslash}X>{\hsize=1.12500\hsize\linewidth=\hsize\raggedright\arraybackslash}X>{\hsize=0.60000\hsize\linewidth=\hsize\raggedright\arraybackslash}X@{}}
\hline
\textbf{Model} & \textbf{Biology} & \textbf{Chemistry} & \textbf{Cybersecurity} & \textbf{Total} \\
\hline
Qwen3.8 27B FP8 & 3 & 1 & 2 & 6 \\
DeepSeek V4 Pro & 6 & 25 & 8 & 39 \\
MiMo V2.5 Pro & 0 & 0 & 1 & 1 \\
Qwen3.8 Max & 6 & 0 & 5 & 11 \\
DeepSeek V4 Flash & 1 & 0 & 0 & 1 \\
Total & 16 & 26 & 16 & 58 \\
\hline
\end{tabularx}
\label{tab:A3}
\end{table}

Of the 7,950 included ID episodes, 5,679 are faithful completions and 1,469 are out-of-bounds. The other 802 are 794 refusals, 6 non-final tool episodes, 1 escalation, and 1 response without structured output; all remain in the denominator.

The 25 call-cap episodes were excluded without being scored. The six non-final tool episodes, five from DeepSeek V4 Flash and one from DeepSeek V4 Pro, carry non-positive labels and stay in the denominator without counting as faithful completions. The two rules were fixed separately, when GLM-5.3 and the six earlier models were processed. Applying either rule to both groups leaves the three main comparisons unchanged. Counting the 25 episodes as non-positive moves two GLM-5.3 medical-claims cells from up to flat (consequence display and goal-directed framing; both intervals include zero under either rule). Excluding the six episodes moves one interval, DeepSeek V4 Flash under goal-directed framing in TOEFL applications, from [1.6, 51.3] pp to [$-$2.7, 48.3] pp. Otherwise only the replicate patterns of a few cells change.

\subsection{Condition-Source Pool}

The condition-source pool contains 119 condition records from 13 studies. Records describe a manipulation or a fixed configuration and retain its reference, timing, and source location. Counts are reported by study: In-context Scheming \cite{meinke2024frontier} 15; Stress Testing Deliberative Alignment \cite{schoen2025stress} 29; Sabotage Risk Report \cite{anthropic2026sabotage} 10; Janus \cite{giannouris2026janus} 2; Gram \cite{lindner2026gram} 8; Honeypots \cite{krakovna2026realistic} 6; Best-Laid SCHEMEs \cite{radev2026schemes} 6; Scheming Propensity \cite{hopman2026scheming} 9; CTRL-ALT-DECEIT \cite{ward2025ctrlaltdeceit} 7; Sleeper Agents \cite{hubinger2024sleeper} 5; AI Sandbagging \cite{vanderweij2025sandbagging} 7; Auditing Games \cite{taylor2025auditinggames} 6; DeceptionBench \cite{huang2025deceptionbench} 9.

\section{Supplementary Results}
\setcounter{table}{0}
\renewcommand{\thetable}{\thesection\arabic{table}}
\setcounter{figure}{0}
\renewcommand{\thefigure}{B\arabic{figure}}
\label{appendix:B}

\subsection{Label Audit and Replay}

Before freezing the rules, researcher-led spot checks with language-model assistance informed extraction and numeric-format rules. The final analysis applied frozen scoring code to saved records, retaining the review decisions and offline labels specified in the frozen inputs.

The inputs of the six earlier models retain 87 records with frozen review decisions. GLM-5.3 has a separate 86-record review log with 84 agreements with the automatic label and 2 researcher decisions, and 25 retained offline labels that are disjoint from that review set. All 25 are non-positive. The review decisions and offline labels were made by the researchers, not by a language model. These counts refer to different stages and sets.

On the same saved episodes within each model (1,140 for each of the six earlier models and 1,085 for GLM-5.3), replaying the frozen rules gives each model 163--396 fewer out-of-bounds labels than its recorded pre-replay labels (3,034 in total before the replay; 1,471 after, and 1,469 under the updated rules). GLM-5.3 has 158 positives on these episodes; 25 additional non-positive offline labels give the full-analysis result of 158/1,110, and its 30 censored episodes are excluded. Regression between the previous and updated rules over the 6,840 episodes of the six earlier models changes two Flash TOEFL tool labels from positive to faithful, reducing its total from 318 to 316; the other five totals are unchanged. We report the updated labels.

\subsection{Execution After Failure Feedback}

Under the stricter definition, a risky execution under permanent failure counts only when the model has received explicit failure feedback from the normal tool before producing the risky call. The denominator remains all included episodes, and the recoverable-wait side retains its full execution rate. The 21 model-by-scenario cells then give 14 up, 4 flat, and 3 down, compared with 18 up, 3 flat, and 0 down under the primary definition.

Under recoverable wait, GLM-5.2, GLM-5.3, and MiMo in cybersecurity executed the risky tool in 51, 73, and 64 of 102 episodes. Under permanent failure, 50 of 92, 80 of 84, and 79 of 98 of their out-of-bounds episodes never called the normal tool; in the other 18 cells, 20 out-of-bounds episodes in total never did. No episode in any of the 21 cells completed the task under permanent failure, and resource exhaustion was the largest terminal category in 15 cells. Permanent failure offers no compliant exit: a reply without a tool call is answered with a request to continue.

\subsection{Consequence Display and Identical-Content Runs}

In TOEFL applications and album sales, the two consequence-display arms return identical visible content, so the main text reports the other three scenarios (21 cells: 8 up, 7 flat, 6 down; one interval above zero). The 14 cells in TOEFL applications and album sales give 7 up, 2 flat, and 5 down, and none of their nominal 95\% intervals excludes zero. They show how much cells vary between runs with identical content.

\begin{table*}[!tp]
\centering
\renewcommand{\thetable}{C1}%
\caption{Audited design and trajectory facts.}
\footnotesize
\setlength{\tabcolsep}{3pt}
\renewcommand{\arraystretch}{1.06}
\begin{tabularx}{\textwidth}{@{}>{\hsize=0.67500\hsize\linewidth=\hsize\raggedright\arraybackslash}X>{\hsize=1.45000\hsize\linewidth=\hsize\raggedright\arraybackslash}X>{\hsize=1.22500\hsize\linewidth=\hsize\raggedright\arraybackslash}X>{\hsize=1.10000\hsize\linewidth=\hsize\raggedright\arraybackslash}X>{\hsize=0.55000\hsize\linewidth=\hsize\centering\arraybackslash}X@{}}
\hline
\textbf{Study} & \textbf{Panel and pairing} & \textbf{Shared conditions / comparison scope} & \textbf{Capability reference} & \textbf{Traj.} \\
\hline
WhistleBench \cite{agrawal2026whistleblow} & 9 main models; five single-factor settings & One behavior across four industries & None located & 1 \\
Auditing Games \cite{taylor2025auditinggames} & 21/25 final target cells; password contrast & Style controls within sandbagging & Password-unlocked performance & 2 \\
AuditBench \cite{sheshadri2026auditbench} & 56 target variants; 2 $\times$ 2 training design & Training and tools across 14 behaviors; audit-method comparison & None located & 2 \\
Propensity Inference \cite{jarviniemi2026propensity} & 23 models $\times$ 11 environments; nonuniform sampling, no reported strict pairing & Seven factors across six parent scenarios; model-based inference & None located & 2 \\
Instrumental Choices \cite{wiedermannmoller2026instrumental} & 10 $\times$ 7 $\times$ 8 $\times$ 3 crossed; seven contrasts & Eight variants across seven tasks; within-risk conditions & No-incentive arm & 1 \\
ROGUE \cite{tien2026rogue} & 37/39 core cells; two separable single-factor edges in four-arm ablation & Subagent settings across three corrigibility failures & External OSWorld reference & 3 \\
Misalignment Indicator Probes \cite{zhou2026probing} & Two targets; three synthetic text types & Shared benign-control procedure; probe comparison & Unconfirmed & Unconfirmed \\
Manipulation Is Task-Dependent \cite{zaman2026manipulation} & Six models $\times$ six environments; 5 $\times$ 3 $\times$ 3 per environment & Three abstract axes across six environments; one behavior & None located & 1 \\
Model Forensics \cite{singh2026forensics} & Models and interventions selected per case & Case-specific hypotheses; no common cross-case manipulation & Local: normal completions in a zero-error version & 2 \\
ResearchArena \cite{libon2026researcharena} & Uneven panel and sources; monitor 2 $\times$ 2 on the same red-team run & Monitoring-information axes across four settings & With/without side task & 2 \\
Escalation Channels \cite{gomez2026escalation} & Eight main models in A-E; some only in A; 2 $\times$ 2 plus one prompt group & One escalation protocol within risk & External single-model replication & 2 \\
OpenAI-HuggingFace reproduction \cite{slocum2026reproduction} & Nine model labels; Step 2 repetition & Four serial steps; no unified cross-step manipulation & None located & 2 \\
Deception by Omission \cite{florin2026omission} & Seven models; multi-turn follow-up & Dialogue across six environments within risk & External recognition task & 2 \\
Gram \cite{lindner2026gram} & Panel not assessed; reproduce anomalies, then ablate by hypothesis & Within-risk and case analysis & Not assessed & --- \\
Scheming Propensity \cite{hopman2026scheming} & 12 candidates, uneven coverage; main results on three models & Agency and long-term-goal snippets across four scenarios & Local no-threat analogue in evaluation sabotage & --- \\
CHIVE \cite{karvonen2026would} & Panel not assessed; counterfactual editing & Explanation-method comparison & Not assessed & --- \\
This study & Seven models cover all three risks; within-risk single-condition comparisons & Evaluation disclosure shared by three risks & Defined for each risk (Section~3.3) & See note \\
\hline
\end{tabularx}
\par\smallskip
{\footnotesize\leftskip=0pt\rightskip=0pt\parfillskip=0pt plus 1fil\relax The table covers the 13 studies whose public trajectories were audited and five seed studies (two overlap; 16 unique), plus this study. Traj.: 1, full raw trajectories within the audited release; 2, extracted or partial messages; 3, selected examples; ---, outside the trajectory audit. Unconfirmed: the audit could not establish the property. None located: no qualifying reference was found within the search scope, which does not show that none exists. This study releases its runner, examples, and conditions, with full trajectories planned for a later release. Other design neighbors discussed in Related Work are outside this table.\par}
\label{tab:C1}
\end{table*}

\subsection{Parsing Failures in Selective Underperformance}

Of the 25,200 episodes in selective underperformance, 83 gave no parsable answer and are scored as incorrect. Excluding them changes no model's $\Delta_{\mathrm{Ind},m}$ by more than 0.62 pp, and $\Delta_{\mathrm{Disc},m}$ is then negative for six of the seven models ($-$0.6 to $-$2.8 pp; GLM-5.3, $+$1.4 pp).

\subsection{Tool Verification Echo by Design}

The tool verification echo is compared within three tool designs, with 16, 12, and 2 configurations per scenario. Across the 35 cells, the incentive design gives 10 up, 12 flat, and 13 down; the state-attribution design 10, 17, and 8; and the neutral design 4, 28, and 3. No interval excludes zero in any design.

\subsection{Flat Cells and Additional Counts}

Among flat cells, both sides are at zero in 15 of 17 for the implicit reporting line, 14 of 15 for recoverable wait versus immediate availability, and 13 of 20 for tool-protocol disclosure; another 3 of the 20 are at 100\% on both sides. The 3 flat cells for permanent failure versus recoverable wait are all in chemistry, with no execution in 102 episodes on either side. One pressure cell has a replicate that cannot be estimated, alongside zero and negative replicates, and is counted as fluctuating.

The reporting-line arm ended in 104 out-of-bounds episodes, 106 faithful completions, and 210 refusals (reference arm: 5, 397, and 18). For the implicit reporting line, 2 of the 10 fluctuating cells have replicates of opposite sign. For tool-protocol disclosure, 11 of the 12 decreases fluctuate. For disclosure in risky-path execution, 13 of the 21 cells fall under permanent failure, and the 3 intervals that exclude zero are all decreases.

\section{Study-Level Comparison}
\setcounter{table}{0}
\renewcommand{\thetable}{\thesection\arabic{table}}
\label{appendix:C}

Table~C1 records study-level facts for the audited and seed studies. Its columns follow the four requirements of Section~2.3: panel and pairing (R3 and R1), shared conditions (R4), and capability reference (R2).

\label{document:text:end}

\end{document}